\documentclass[letterpaper, 10 pt, conference]{ieeeconf}  

\IEEEoverridecommandlockouts                              

\usepackage{graphicx}
\usepackage{float}
\usepackage{amsmath}
\usepackage{booktabs}

\usepackage{lipsum} 
\usepackage{booktabs} 
\usepackage{multicol} 
\usepackage{float} 
\usepackage{adjustbox} 
\usepackage{multirow} 
\usepackage{amssymb}

\usepackage{threeparttable}

\usepackage{biblatex}
\title{
{\Large \textbf{Iterative Volume Fusion for Asymmetric Stereo Matching}}
}
\author{Yuanting Gao$^{\dag}$\thanks{\dag Tsinghua Shenzhen International Graduate School, Shenzhen, China.} 
and 
Linghao Shen$^{\ddag,*}$
\thanks{\ddag Research and Development Center, Sony (China) Ltd.}
\thanks{*Corresponding author}
}

\begin{document}

\maketitle
\thispagestyle{empty}
\pagestyle{empty}

\begin{abstract}
Stereo matching is vital in 3D computer vision, with most algorithms assuming symmetric visual properties between binocular visions. However, the rise of asymmetric multi-camera systems (e.g., tele-wide cameras) challenges this assumption and complicates stereo matching. Visual asymmetry disrupts stereo matching by affecting the crucial cost volume computation. To address this, we explore the matching cost distribution of two established cost volume construction methods in asymmetric stereo. We find that each cost volume experiences distinct information distortion, indicating that both should be comprehensively utilized to solve the issue. Based on this, we propose the two-phase Iterative Volume Fusion network for Asymmetric Stereo matching (IVF-AStereo). Initially, the aggregated concatenation volume refines the correlation volume. Subsequently, both volumes are fused to enhance fine details. Our method excels in asymmetric scenarios and shows robust performance against significant visual asymmetry. Extensive comparative experiments on benchmark datasets, along with ablation studies, confirm the effectiveness of our approach in asymmetric stereo with resolution and color degradation.

\end{abstract}

\section{Introduction}
Perceiving environmental depth through images is essential in fields like autonomous driving~\cite{app-autodrive}, robotics~\cite{app-robot}, and 3D reconstruction~\cite{app-3drec}. Stereo matching, which estimates depth by calculating the disparity of a point along the epipolar lines in two stereo camera views, is vital in 3D vision. Learning-based methods~\cite{igev, mc, 2024mocha, xu2024igev++}, utilizing CNNs and neural networks for feature extraction and representation learning, have shown impressive performance in scenarios where both images are high-quality and symmetric.

\begin{figure}[t!]
      \centering
      \includegraphics[width=\linewidth]{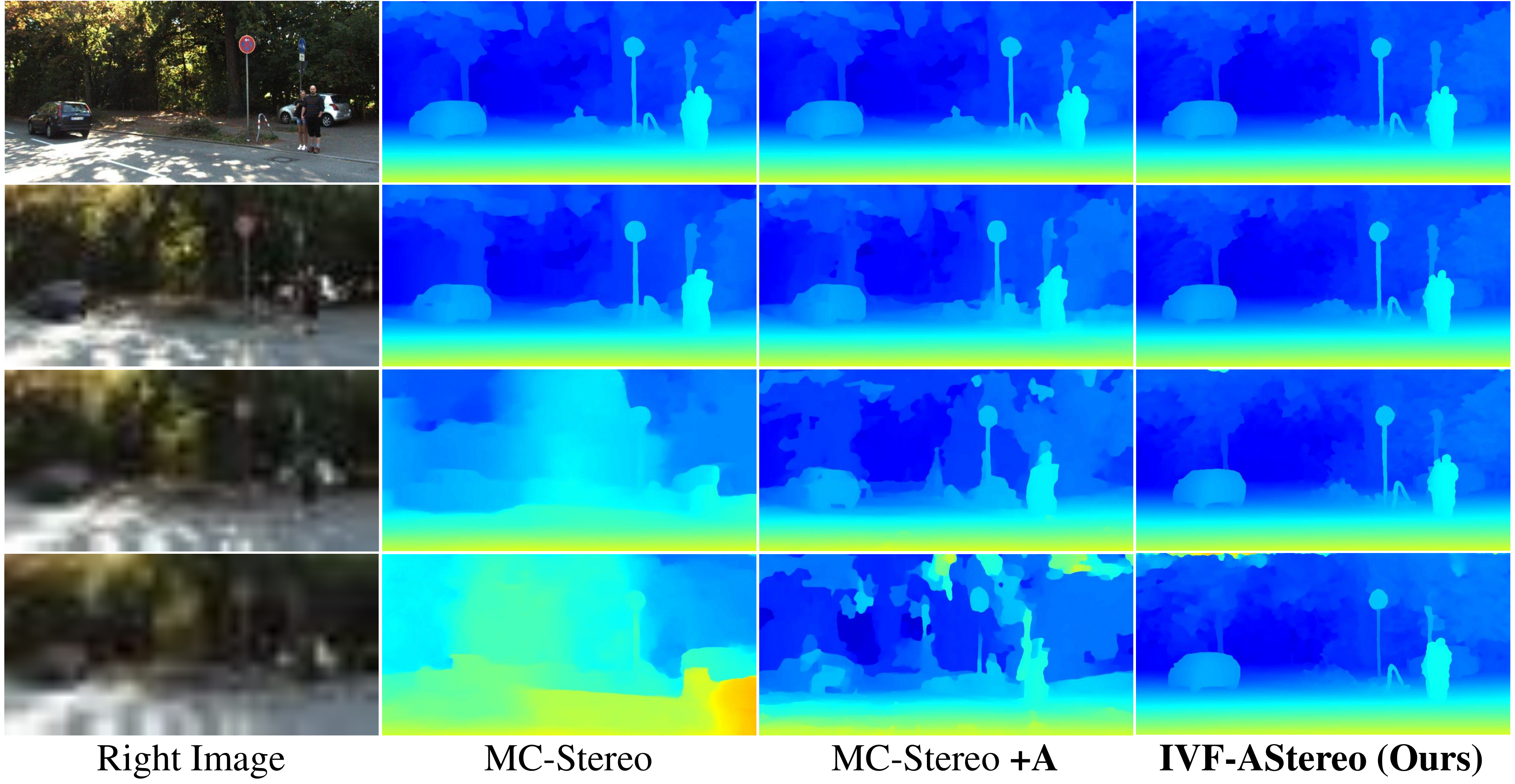}
      \caption{Asymmetric Stereo Matching example with lower right view resolution than left view. Rows show input right views in original size $(H, W)$, $(\frac{H}{2}, \frac{W}{2})$, $(\frac{H}{6}, \frac{W}{6})$, and $(\frac{H}{8}, \frac{W}{8})$ with corresponding estimated disparities. Representative existing method MC-Stereo significantly degrades when right view is downsampled, but adding the proposed mitigation (MC-Stereo +\textbf{A}) substantially restores performance. Proposed method achieves more accurate disparity estimation regardless of asymmetric resolution.} \label{fig1}
\end{figure}

However, to reduce expense and get more thorough information, asymmetric camera system are commonly adopted, such as multi-focal camera setups on robots or autonomous vehicles and tele-wide cameras, in which different cameras may have different resolutions, color depths, lens blur, imaging modalities, etc.~\cite{Dual_Lens, Degradation_Agnostic_stereo}. Additionally, promising but special color filter solutions (e.g., RCCB and RCCC) can also introduce large color asymmetry~\cite{Brucker_2024_CVPR}. In such scenarios, methods designed for symmetric camera system may degrade substantially when the one view's quality deteriorates (See Figure~\ref{fig1}). Therefore, a significant challenge is how to directly utilize cameras designed for different tasks for stereo matching to obtain environmental depth, without the need for adding a symmetrical stereo camera system. To address this challenge, some existing works~\cite{Degradation_Agnostic_stereo,Liu_2020_CVPR,asymmetric_stereo_selfsim} have attempted to improve the performance of models in asymmetric settings by either repairing degraded images beforehand or proposing degradation-agnostic consistency and utilizing unsupervised learning paradigms. However, few efforts tackle this issue by leveraging the inherent capability of the model.

This study identifies that the distortion of cost volume can be the primary factor contributing to performance degradation. The cost volume, which critically captures similarities between corresponding pixels in stereo views, is typically constructed via feature correlation~\cite{GwcStereo,CoEx-stereo,raft} (correlation volume) or disparity-based feature concatenation~\cite{mc,GCNet,psm,ACV_Stereo} (concatenation volume). However, as Figure~\ref{fig2} shows, asymmetric visual properties can cause the matching cost distribution in the correlation volume to deviate significantly from the ground truth, negatively impacting stereo matching. Nevertheless, eliminating the correlation volume also results in the loss of fine-detail information. Conversely, the concatenation volume, while more robust to visual asymmetry due to complex operations, may lack fine-detail information owing to downsampling in 3D convolutions. Thus, we posit that the key solution to asymmetric stereo matching lies in the comprehensive utilization of distorted cost volumes.

Based on this finding, we propose the two-phase Iterative Volume Fusion network for asymmetric stereo matching (\textbf{IVF-AStereo}). Initially, we reconfigure the correlation volume computation to mitigate asymmetric stereo's adverse effects. We downgrade the high-quality view to generate symmetric features for computing the correlation volume and employ multi-peak lookup~\cite{mc} to handle the flat and multi-peak matching cost distribution. Subsequently, we iteratively estimate the disparity using two complementary GRU branches in a two-phase scheme. In the first phase, the robust concatenation volume refines the degraded correlation volume. In the second phase, we enhance fine-detail estimation by integrating the refined correlation volume with the concatenation volume. By optimizing cost volume computation and strategically fusing information within two GRU branches, we effectively leverage the model's capability to address challenges in asymmetric stereo matching.

Our contributions are summarized as follows:
\begin{itemize}

\item To analyze the fundamental cause of deterioration in stereo matching with asymmetric visual properties, we quantify its adverse effects through the distortion of matching cost distribution. This quantitative study inspires us to propose the strategy for mitigating the distortion and enhance information utilization.

\item We present IVF-AStereo, which utilizes two cost volumes and achieves a superior performance. The model is trained end-to-end with information flow differing in two phases to obtain a more precise depth estimation. To the best of our knowledge, this study uniquely addresses stereo matching in asymmetric scenarios by leveraging the inherent capabilities of the model without using multistage training or additional image restorations.  

\item We conduct comprehensive experiments on Scene Flow~\cite{scenflow}, KITTI~\cite{kitti2012, kitti2015}, and Middlebury~2014~\cite{middlebury} datasets, focusing on resolution and color asymmetry to demonstrate our model's efficacy in asymmetric stereo matching. Additionally, we conduct ablation studies to validate the effectiveness of our proposed approach.

\end{itemize}

\section{Related Work}

\subsection{Cost Volume in Stereo Matching}
It is crucial in stereo matching to construct an informative cost volume that can incorporate the matching cost along the epipolar line. GCNet~\cite{GCNet} initially utilizes a 3D encoder-decoder to regularize a 4D cost volume, which is constructed by concatenating paired features across disparity levels. PSMNet~\cite{psm} proposes a stacked hourglass 3D CNN to better regularize the concatenation volume. GwcNet~\cite{GwcStereo} propose to construct the cost volume by group-wise correlation and ACVNet~\cite{ACV_Stereo} propose an attention concatenation volume. IGEV-Stereo~\cite{igev} builds a combined geometry encoding volume to encodes geometry and context information as well as local details. 
However, we observe the cost volume can be degraded by asymmetric visual properties, resulting in substantial stereo matching performance decline. Nevertheless, few studies have focused on the impact of asymmetric stereo matching on the cost volume and its mitigation.

 \begin{figure}[thpb]
      \centering
      \includegraphics[width=\linewidth]{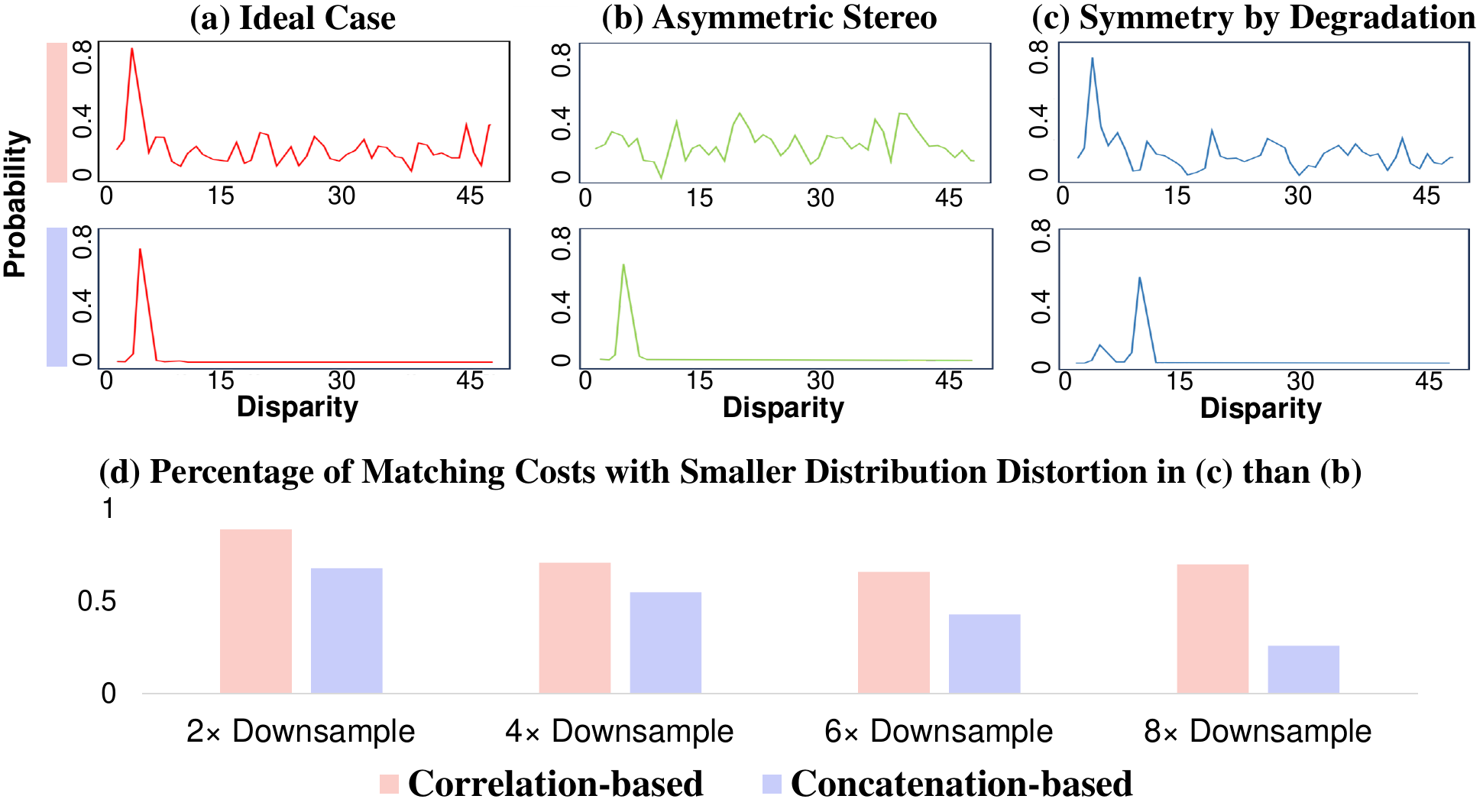}
      \caption{Empirical study of matching cost distribution distortion with asymmetric stereo on KITTI dataset for three cases: \textbf{(a) Ideal case} with two high quality views. \textbf{(b) Asymmetric stereo} with one high quality view and one distorted view. \textbf{(c) Symmetry by degrading} the high quality view in (b). Demonstrative distributions of correlation volume (top) and concatenation volume (bottom) are given correspondingly. Comparing the distributions in (b) and (c) against the ideal case, correlation-based cost in (c) is less distorted, while concatenation-based cost in (b) is more desirable. \textbf{(d)} summarizes the percentage of matching costs that have smaller distribution distortion (details in Section \ref{sec:emp-study}) in (c) than (b) for increasing level of downsample distortion. It is observed that correlation volume consistently favors case (c) across different distortion levels, whereas concatenation volume exhibits progressively diminished performance.} \label{fig2}
\end{figure}

\subsection{Iterative Disparity Optimization}
RAFT-Stereo~\cite{raft} first introduces multi-level convolutional GRUs to efficiently propagate information across the image and optimize disparity in a coarse-to-fine manner. CREStereo~\cite{cre_stereo} refines disparity iteratively using recurrent update module, which apply adaptive group correlation layer to tackle more practical stereo matching. IGEV-Stereo~\cite{igev} indexes combined geometry encoding volume through iterative refinement, achieving remarkable performance and cross-dataset generalization capability. MC-Stereo~\cite{mc} introduces a multi-peak lookup strategy and a cascade search range to effectively index additional information within the volume. This approach is critical for addressing the multi-peak distribution in the cost volume. 
While the iterative optimization can be effective in asymmetric stereo matching, existing methods, which are designed for symmetric matching, still perform suboptimally even the cost volume mitigation method is applied. Nonetheless, by redesigning a two-phase iterative fusion process for asymmetric stereo matching, we can achieve outstanding result.

\subsection{Asymmetric Stereo Matching}
Several studies have explored stereo depth estimation under asymmetric settings where the image of one view is distorted. Liu et al.~\cite{Liu_2020_CVPR} employs guided view synthesis to restore the distorted view before stereo matching. To address photometric and radiometric difference in the loss computation, some methods employ mutual information~\cite{egnal2000mutual} or Adaptive Normalized Cross-Correlation~\cite{Adaptive—correlation} as the consistency measure. Recently, Chen et al.~\cite{Degradation_Agnostic_stereo} proposes a degradation-agnostic feature-metric consistency to address the failure of photometric consistency and learns the feature space unsupervisely. Building on this, Taeyong Song et al.~\cite{asymmetric_stereo_selfsim} presents the spatially-adaptive self-similarity to further enhance feature consistency in unsupervised asymmetric stereo matching. Existing works prioritize restoration or distortion-agnostic feature extraction for imposing view consistency. However, our study (Figure~\ref{fig2}) suggests that more inherent solution could reside in resolving distorted cost volumes. To overcome this challenge, we focus on fusing complementary cost volumes for asymmetric stereo matching. 

\begin{figure*}
\centering
\includegraphics[width=\textwidth]{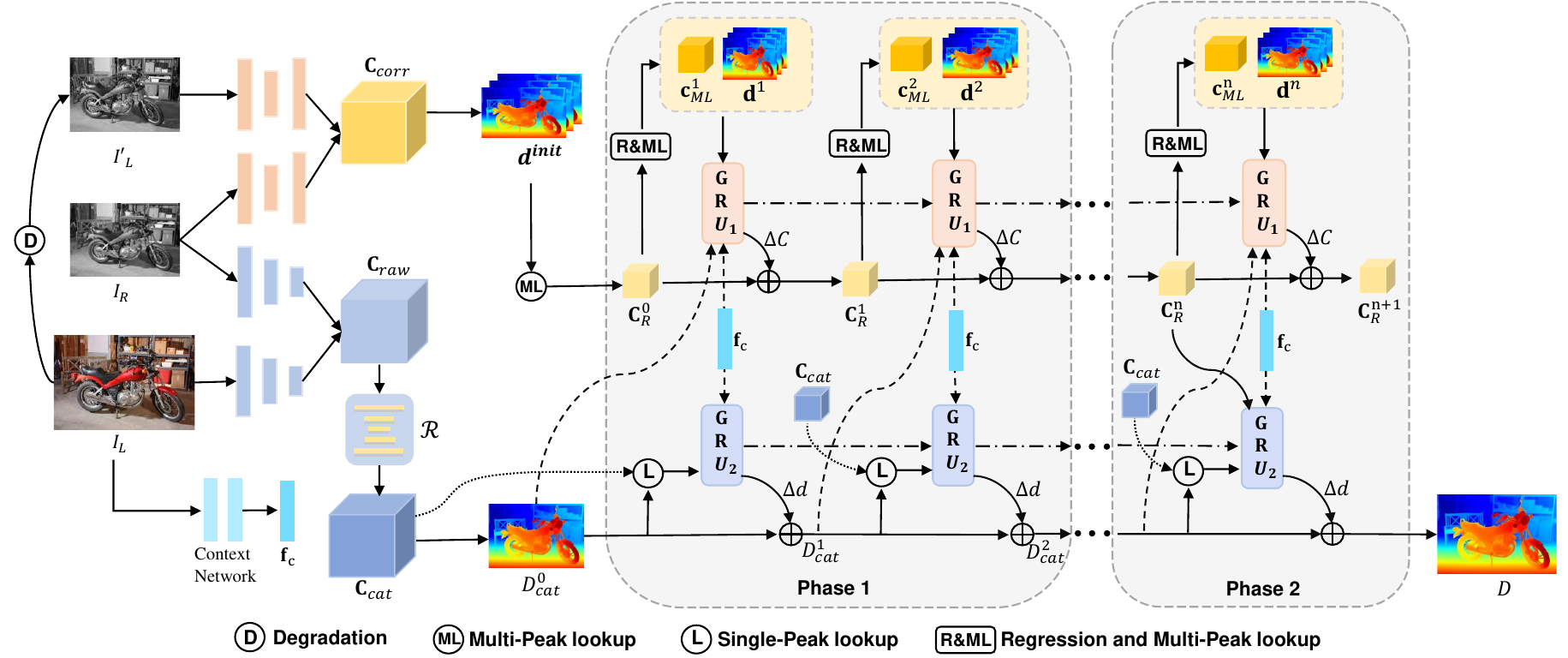}
\caption{The workflow of IVF-AStereo. First, given \(I_L\) and \(I_R\), two parallel feature extractors are employed to tackle the asymmetric property (Section \ref{sec:feat-ext}). Besides, in conjunction with the deliberately degraded $I'_L$, two cost volumes are constructed to utilize the unbalanced information (Section \ref{sec:volume}). Then, two GRU branches iteratively refine the local correlation volume ($\text{GRU}_1$) and disparity ($\text{GRU}_2$) in a two-phase approach for disparity optimization (Section \ref{sec:gru}). Ultimately, $\text{GRU}_2$ produces the final disparity estimation.}
\label{fig:workflow}
\end{figure*}

\section{Method}
To address challenges in asymmetric stereo, in this section, we first quantitatively investigate its impact through the distortion of matching cost distribution. Then, we introduce our mitigation in feature extractor and the cost volume constructions under asymmetric stereo conditions. Moreover, we describe the structure of the two complementary GRU branches and the two-phase information flow (see Figure~\ref{fig:workflow}). Without loss of generality, let left view image \( I_L \) be the high quality image with resolution $H \times W$ and right view image $I_R$ with degradation and lower resolution $\frac{H}{k} \times \frac{W}{k}$.

\subsection{Distortion of Matching Cost Distribution} \label{sec:emp-study}
Matching cost is a critical component in stereo matching, serving as an intermediary between input visual features and disparity correspondence. Hence, to quantitatively study the effect of asymmetric stereo, we focus on the distortion of the matching cost distribution. Utilizing a pretrained stereo matching model IGEV-Stereo~\cite{igev}, we compute the matching cost with both symmetric ($c_\text{ideal}$) and asymmetric ($c_\text{asym}$) inputs. Then, the distribution distortion ($\text{DD}$) is defined via Kullback-Leibler divergence as:

\begin{equation}
\text{DD}(c_\text{asym}) = \text{D}_{\text{KL}} \left( 
\text{softmax}(c_\text{asym}) \| 
\text{softmax}(c_\text{ideal}) \right).
\end{equation}

Figure~\ref{fig2} illustrates the $\text{DD}$ of correlation-based and concatenation-based matching cost. We observe that the concatenation-based matching cost performs effectively under asymmetric stereo. Conversely, for correlation-based matching cost, it may be necessary to degrade the high-quality view. Due to the averaging effect of downsampling, the probability density is less concentrated and exhibits a tendency to produce multiple peaks in the distribution. These observations inspire our design in the subsequent sections.

\begin{figure*}[t]
\centering
\includegraphics[width=\textwidth]{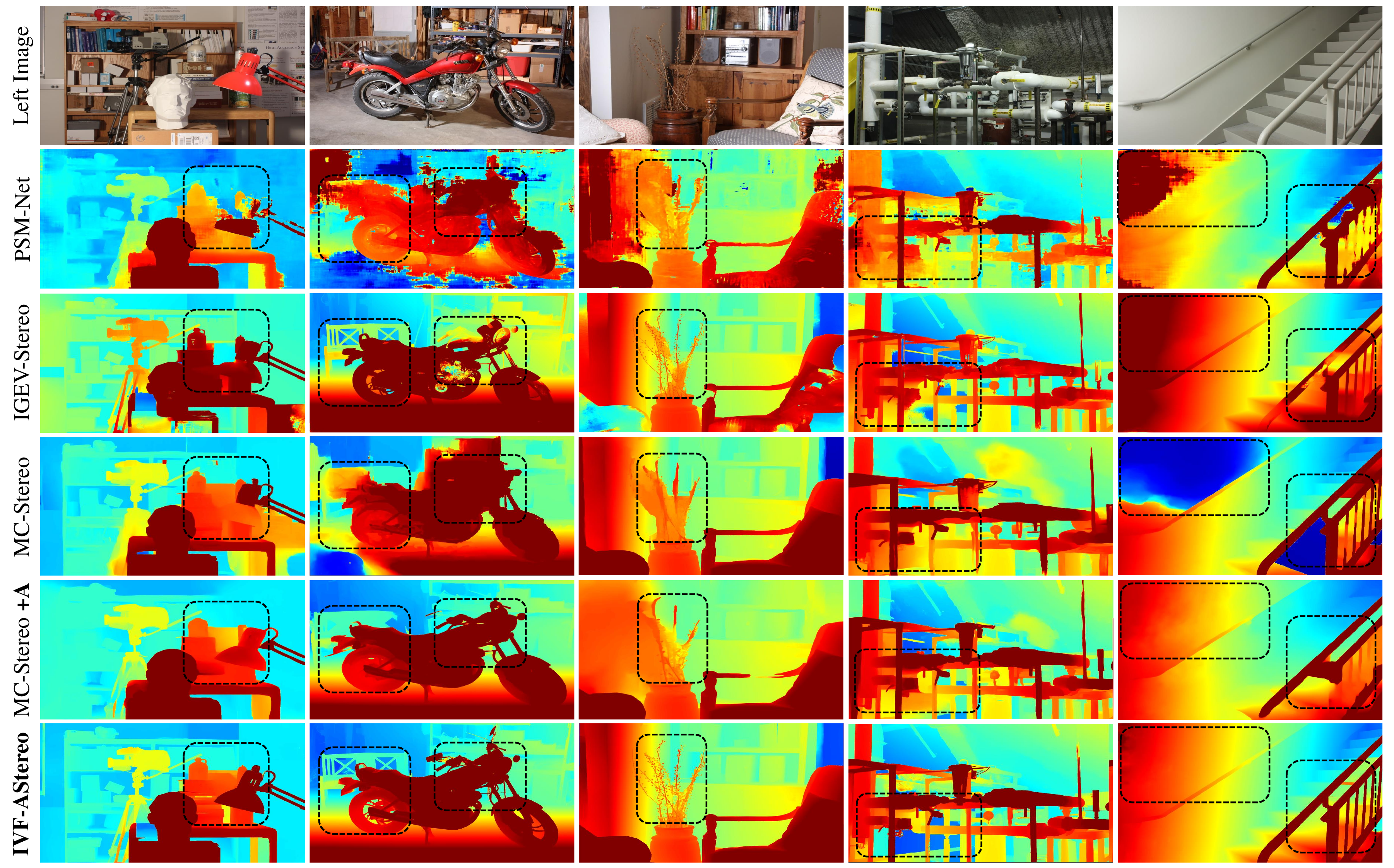}
\caption{Visual examples of zero-shot depth estimation on Middlebury~2014~\cite{middlebury}. Our IVF-AStereo performs well in both overall depth and fine details with asymmetric stereo. Applying our mitigation (\textbf{+A}) to MC-Stereo can also yields notable improvements.}
\label{fig:example}
\end{figure*}

\subsection{Mitigation in Feature Extractor} \label{sec:feat-ext}
Though using a shared feature extractor for all cost volume constructions can be common in stereo matching~\cite{igev, raft}, employing separate feature extractors for correlation and concatenation volumes may be necessary in asymmetric stereo due to their distinct effects on cost volumes. For correlation volume, features \( \mathbf{f}^{cor}_{l}, \mathbf{f}^{cor}_{r}\in R^{C \times \frac{H}{4} \times \frac{W}{4}} \) are extracted using ConvNext pre-trained on ImageNet~\cite{ImageNet} with U-Net style upsampling, following MC-Stereo~\cite{mc}. For concatenation volume, multi-scale features $ \mathbf{f}^{cat}_{l,i}, \mathbf{f}^{cat}_{r,i} \in R^{C_i \times \frac{H}{i} \times \frac{W}{i}}$ for $i=4,8,16,32$ are extracted using MobileNetV2 pretrained on ImageNet, following IGEV-Stereo~\cite{igev}. $\mathbf{f}^{cat}_{l,4}$ and $\mathbf{f}^{cat}_{r,4}$ construct the concatenation volume, while other scales guide the 3D convolutional regularization network. Moreover, we extract context features $\mathbf{f}_c$ from high quality view for guiding depth estimation. Akin to RAFT-Stereo~\cite{raft}, residual blocks and downsampling layers are used to produce features in \( \frac{1}{4} \), \( \frac{1}{8} \), and \( \frac{1}{16} \) of the input size, which are then inserted into the ConvGRU at the corresponding scale.

\subsection{Mitigation in Cost Volume Construction} \label{sec:volume}
Unlike cost volumes in symmetric stereo that are generally constructed with the same input, different cost volumes in asymmetric stereo can favor different input properties. As depicted in Figure~\ref{fig2}, correlation volumes highly rely on the symmetric quality of two views. However, concatenation volumes are more robust to quality asymmetry and better leverage the high-quality view. Thus, we tailor cost volume construction methods accordingly.

\subsubsection{Correlation Volume}
Given \( I_L \) and \( I_R \), to facilitate symmetric quality, \( I_L \) is deliberately degraded into \( I_L' \) according to the degradation of $I_R$. Notably, degradation is simpler than restoration and introduce less bias.
Obtaining \( \mathbf{f}^{cor}_{l} \) and \( \mathbf{f}^{cor}_{r} \) from \( I_L' \) and \( I_R \) respectively, the correlation volume is thenconstructed as:
\begin{equation}
\mathbf{C}_{corr}(d,x,y) = \left\langle \mathbf{f}^{cor}_{l}(x,y),\mathbf{f}^{cor}_{r}(x-d,y) \right\rangle ,
\end{equation}
where \( \langle \cdot, \cdot \rangle \) represents the inner product, and \( d \) denotes the searching disparity in $\left[0,D_{max}/4 \right]$ ($D_{max}=192$).

\subsubsection{Concatenation Volume}
To reduce computational cost while preserving useful information, we employ Group-Wise concatenation~\cite{GwcStereo} to obtain \( \mathbf{f}^{cat}_{l,i} \) and \( \mathbf{f}^{cat}_{r,i} \) from $I_L$ and upsampled $I_R$. The channels of \( \mathbf{f}^{cat}_{l,4} \) and \( \mathbf{f}^{cat}_{r,4} \) are partitioned into $G$ groups, and dot products are computed within each group to produce the raw concatenated volume:
\begin{equation}
\mathbf{C}_{raw}(g,d,x,y) = \frac{1}{G} \left\langle \mathbf{f}^{cat}_{l,4,g}(x,y),\mathbf{f}^{cat}_{r,4,g}(x-d,y) \right\rangle ,
\end{equation}
where ${f}^{cat}_{\cdot,4,g}$ denotes the $g$-th channel group. A lightweight 3D regularization network $\mathcal{R}$ is applied on $\mathbf{C}_{raw}$ to produce the final concatenation volume~\cite{igev}
\begin{equation}
\mathbf{C}_{cat} = \mathcal{R}\left(\mathbf{C}_{raw}, \mathbf{f}^{cat}_{l,i=8,16,32}\right)
\end{equation}
augmented by multi-scale $\mathbf{f}^{cat}_{l,i}$ from the high-quality view.

\subsection{Two-phase Iterative Volume Fusion} \label{sec:gru}
Iterative depth optimization with a ConvGRU has shown effectiveness in symmetric stereo matching~\cite{raft,cre_stereo,igev}. Yet, employing two complementary ConvGRUs can be more efficient for tackling the intricacies in asymmetric stereo.

\subsubsection{Local Correlation Volume Refinement}
Given the multi-peak characteristic of the correlation-based matching cost distribution in asymmetric stereo, following~\cite{mc}, a ConvGRU ($\text{GRU}_1$) is employed to iteratively refine the correlation volume $\mathbf{C}^{i}_{R}$ by extracting the top-$K$ similarity peaks from the previous iteration's volume $\mathbf{C}^{i-1}_{R}$ with a multi-peak lookup~\cite{mc}. In addition, the near-peak cost volume $\mathbf{c}_{k}$ is indexed from $\mathbf{C}_{cor}$ centered at the $k$-th peak's disparity $d_{k}$ with a decreasing search range $r_i$ across iterations:
\begin{equation*}
\mathbf{c}_{k} = \small\left[ \mathbf{C}_{cor}(d_{k} - r_{i}), \ldots , \mathbf{C}_{cor}(d_{k}), \ldots, \mathbf{C}_{cor}(d_{k} + r_{i}) \right],
\end{equation*}
where $[\cdot]$ denotes feature concatenation.

Finally, $\mathbf{C}^{i-1}_{R}$ is updated by $\text{GRU}_1$ with the multi-peak local cost volume $\mathbf{c}_{ML} = \left[ \mathbf{c}_{1}, \ldots \mathbf{c}_K \right]$,
$\mathbf{d} = \left[d_{1}, \ldots, d_{K} \right]$, context features
$\mathbf{f}_c$, and the latest disparity map $D^{i}_{cat}$:
\begin{equation}
    \Delta C = \text{GRU}_1\left( \left[\mathbf{c}_{ML}, \mathbf{d}, D^{i}_{cat}\right], \mathbf{f}_c \right),
\end{equation}
\begin{equation}
    \mathbf{C}^{i}_{R}= \mathbf{C}^{i-1}_{R}+ \Delta C.
\end{equation}

In the first iteration, $\mathbf{d}$ and $\mathbf{c}_{ML}$ are retrieved from $\mathbf{C}_{cor}$ rather than $\mathbf{C}^{i-1}_{R}$ and set $\mathbf{C}^{0}_{R} = \mathbf{c}_{ML}$. The $\mathbf{f}_c$ is used to initialize the hidden state of $\text{GRU}_1$.

\subsubsection{Iterative Disparity Optimization with Two-phase Fusion} \label{sec:fusion}
The concatenation volume empirically shows greater robustness against asymmetric stereo compared to the correlation volume. Hence, we utilize another ConvGRU ($\text{GRU}_2$) to proceed the concatenation volume for initial optimization, while $\text{GRU}_1$'s information is incorporated subsequently.

First, the initial disparity $D^0_{cat}$ is given by:
\begin{equation}
    D^0_{cat} = \sum^{D_{max}}_{d=0} d \cdot \text{softmax} \left( \mathbf{C}_{cat}(d) \right).
\end{equation}
Then, for the $i$-th iteration, with the search range $r$, the single-peak local matching cost $\mathbf{c}_{L}$ is retrieved as
\begin{equation*}
    \mathbf{c}_{L} = \small\left[ \mathbf{C}_{cat}(D^{i-1}_{cat} - r),\ldots, 
    \mathbf{C}_{cat}(D^{i-1}_{cat} + r) \right].
\end{equation*}
Finally, the disparity $D^{i-1}_{cat}$ is updated by $\text{GRU}_2$:
\begin{equation}
    \Delta d=\text{GRU}_2\left( \left[\mathbf{c}_{L}, \mathbf{C}'^{i-1}_{R}, 
    D^{i-1}_{cat}\right], \mathbf{f}_c \right),
\end{equation}
\begin{equation}
    D^{i}_{cat}=D^{i-1}_{cat}+ \Delta d.
\end{equation}
Here, $\mathbf{C}'^{i-1}_{R}$ denotes $\text{GRU}_1$'s information. Incorporating the correlation volume presents challenges due to its initial distortion. To tackle this issue, we adopt a two-phase scheme:
\begin{equation}
    \mathbf{C}'^{i-1}_{R}=\left\{
	\begin{aligned}
	 \mathbf{0} \quad (\text{Phase}~1)\\
     \mathbf{C}^{i-1}_{R} \quad (\text{Phase}~2)\\
	\end{aligned}
	\right.
\end{equation}
Essentially, this two-phase scheme reduces the variance from initially distorted $\mathbf{C}^{i-1}_{R}$ but allows the complementary fusion of refined information afterwards. The effectiveness of this two-phase scheme is supported by the ablation study.

\begin{table}[tbp]
\centering
\caption{Evaluation on the SceneFlow (measured in D1~/~EPE), smaller value indicates better performance.} \label{tab:sceneflow}
\begin{adjustbox}{width=0.9\linewidth} 
\renewcommand*{\arraystretch}{1.15}
\begin{threeparttable}
\begin{tabular}{@{}lcccc@{}}
\hline
\multicolumn{1}{c}{\multirow{2}{*}{\textbf{Method}}} & \multicolumn{4}{c}{\textbf{Right View Downsampling}}                   \\
\cmidrule(l){2-5}
 &
  $ 1 $ &
  $ 2 $ &
  $ 4 $ &
  $ 6 $ \\ \hline
  
PSMnet~\cite{psm}                                    & 3.27~/~0.95             & 33.38~/~6.52             & 50.37~/~9.94 & 62.88~/~12.65 \\
CoEx~\cite{CoEx-stereo}                              & 2.42~/~0.67             & 13.32~/~2.68             & 27.75~/~5.63 & 47.44~/~9.99  \\
FAD-Net~\cite{fadnet}                                & 2.42~/~0.65             & 15.59~/~2.76             & 31.48~/~5.35 & 45.77~/~7.79  \\
CasNet~\cite{CasNet}                                 & 2.62~/~0.67             & 25.13~/~3.68             & 36.29~/~5.20 & 48.94~/~7.02  \\
MC-Stereo~\cite{mc}                                  & 5.06~/~0.45             & ~5.95~/~0.52             & ~9.19~/~0.73 & 17.47~/~1.33  \\
MC-Stereo~+\textbf{A}                                & 5.06~/~\underline{0.45} & ~5.51~/~\underline{0.52} & ~8.86~/~0.73 & 17.21~/~1.31  \\
IGEV-Stereo~\cite{igev}                              & \underline{2.32}~/~0.46 & ~2.57~/~0.56             & ~3.42~/~0.80 & ~6.72~/~1.62  \\
IGEV++~\cite{xu2024igev++}\tnote{*} &
  \textbf{1.81}~/~\textbf{0.43}\tnote{1} &
  ~\textbf{2.00}~/~\textbf{0.48} &
  ~\textbf{2.45}~/~\underline{0.63} &
  ~\underline{4.02}~/~\underline{0.97} \\
\textbf{Ours} &
  2.48~/~0.54 &
  ~\underline{2.49}~/~0.53 &
  ~\underline{2.62}~/~\textbf{0.53} &
  ~\textbf{3.01}~/~\textbf{0.75} \\ \hline

\hline
\end{tabular}
\begin{tablenotes}
\item[*] Newly proposed after this work submitted.
\item[1] \textbf{Bold} and \underline{underline} denote the best and second-best performance respectively.
\end{tablenotes}
\end{threeparttable}
\end{adjustbox}
\end{table}

\subsection{Loss Function}
We compute the loss between the ground truth disparity $D_{gt}$ and the predicted disparity through $N$ refinement iterations. For $\text{GRU}_1$, the $C^{i}_{R}$ is first regressed into disparity
\begin{equation}
    D^{i}_{cor} = \sum^{K}_{k=1} \sum_{d \in \mathbf{c}_{k}} d \cdot \text{softmax} \left( \mathbf{C}^{i}_{R}(d) \right).
\end{equation}
Then, we calculate the L1 loss according to~\cite{mc}:
\begin{equation}
    \mathcal{L}_{\text{GRU}_1}=\sum^N_{i=1}\gamma^{(N-i)} \left \| D^{i}_{cor} - D_{gt} \right \|,
\end{equation}
For $\text{GRU}_2$, we follow~\cite{igev} to use the Smooth L1 loss on initial disparity and L1 loss on refined disparity:
\begin{equation}
    \mathcal{L}_{\text{GRU}_2} = \small \left\|D^0_{cat}-D_{gt}\right\|_{\text{smooth}} + \sum_{i=1}^{N} \gamma^{(N-i)}
    \left \| D^{i}_{cat} - D_{gt} \right \|.
\end{equation}
We use $\gamma = 0.9$ in both loss. The total loss is defined as:
\begin{equation}
    \mathcal{L_{\text{AStereo}}} = \mathcal{L}_{\text{GRU}_1} + \mathcal{L}_{\text{GRU}_2}
\end{equation}

\begin{table}[t]
  \centering
    \caption{Representative results on the KITTI~2012/2015 test set.} \label{tab:kitti}
  \begin{adjustbox}{width=1\linewidth} 
\renewcommand*{\arraystretch}{1.1}
\begin{threeparttable}
\begin{tabular}{@{}lcccc|ccc@{}}
      \toprule
      \multicolumn{1}{c}{\multirow{2}{*}{\textbf{Method}}} & \multicolumn{4}{c|}{KITTI 2012} & \multicolumn{3}{c}{KITTI 2015} \\
 & 3-noc & 3-all & EPE-noc & \multicolumn{1}{c|}{EPE-all} & D1-bg-All & D1-fg-All & D1-All  \\
      \midrule
PSMnet~\cite{psm} (\textbf{S})\tnote{1} & 1.49 & 1.89 & 0.5 & 0.6 & 1.86 & 4.62 & 2.32 \\
CoEx~\cite{CoEx-stereo} (\textbf{S}) & 1.55 & 1.93 & 0.5 & 0.5 & 1.74 & 3.41 & 2.02 \\
GwcNet~\cite{GwcStereo} (\textbf{S}) & \textbf{1.32} & \textbf{1.70} & 0.5 & \textbf{0.5} & 1.74 & 3.93 & 2.11 \\
\textbf{Ours} (\textbf{A})\tnote{2} & 1.52 & 1.87 & \textbf{0.5} & 0.6 & \textbf{1.62}  & \textbf{2.56} & \textbf{1.78} \\
      \bottomrule
    \end{tabular}
    \begin{tablenotes}
\item[1] \textbf{S}: Symmetric input with no degradation.
\item[2] \textbf{A}: Asymmetric input with grayscale and quarterly downsampled right view images.
\end{tablenotes}
\end{threeparttable}
  \end{adjustbox}

\end{table}

\section{Experiments}
We conducted experiments on the Scene Flow~\cite{scenflow}, KITTI~\cite{kitti2012, kitti2015}, and Middlebury~2014~\cite{middlebury} datasets, focusing on the resolution and color asymmetry commonly appearing in multi-camera systems. In all datasets, the left views were preserved, whereas the right views underwent color and resolution degradation through RGB-to-grayscale conversion and bilinear downsampling. Unless otherwise specified, quarter downsampling was applied. MC-Stereo relies exclusively on correlation volume. Thus, for MC-Stereo, both default image processing (upsampling) and the proposed method (degrading), denoted as +\textbf{A}, for achieving symmetric input were assessed to demonstrate our approach's effectiveness. Training and evaluation protocols adhered to those in ~\cite{igev, mc} and utilized the OpenStereo~\cite{OpenStereo} framework to compare various methods under different asymmetric settings.

\begin{table}[t]
  \centering
    \caption{Zero-shot evaluation on the Middlebury 2014 (measured in EPE), smaller value indicates better performance.}  \label{tab:middlebury}
  \begin{adjustbox}{width=0.65\linewidth} 
  \begin{threeparttable}
    \begin{tabular}{l cc cc cc cc}
      \toprule
    \multicolumn{1}{c}{\multirow{2}{*}{\textbf{Method}}} & \multicolumn{4}{c}{\textbf{Right View Downsampling}}                   \\
    \cmidrule(l){2-5}
     &
      $ 2 $ &
      $ 4 $ &
      $ 6 $ &
      $ 8 $ \\ \hline
    \midrule
    PSMnet~\cite{psm} & 33.39 & 44.00  & 46.45   & 44.01   \\
    CoEx~\cite{CoEx-stereo}& 23.63  & 39.17  & 44.43  & 45.86 \\
    FAD-Net~\cite{fadnet} & 14.10  & 26.22  & 35.80   & 42.39    \\
    CasNet~\cite{CasNet} & 24.11  & 38.60  & 44.91   & 46.16   \\
    MC-Stereo~\cite{mc} & 4.34  & 5.94    & 9.35   & 17.17   \\
    MC-Stereo +\textbf{A} & 3.50  & 3.73    & \underline{4.47}   & \underline{5.38}     \\
    IGEV-Stereo~\cite{igev}& \underline{1.35} & \underline{2.62}  & 7.40  & 12.57  \\
    IGEV++~\cite{xu2024igev++}\tnote{*} & 2.16 & 2.99  & 5.07  & 8.58  \\
    \textbf{Ours} & \textbf{0.99}  &\textbf{1.34}  & \textbf{2.38} &  \textbf{3.34} \\
    \bottomrule
  \end{tabular}
\begin{tablenotes}
\item[*] Newly proposed after this work submitted.
\end{tablenotes}
\end{threeparttable}
  \end{adjustbox}

\end{table}

\subsection{Evaluation on Scene Flow}
We compared our IVF-AStereo with state-of-the-art methods on the Scene Flow dataset. As shown in Table~\ref{tab:sceneflow}, our method achieved superior result in asymmetric scenarios. Meanwhile, MC-Stereo~\textbf{+A} with degraded but symmetric inputs, outperformed its default configuration, corroborating our observation on the matching cost distortion. Moreover, our method exhibits stronger robustness as the right view resolution degradation (downsampling factor) increases. Notably, the newest IGEV++ with multi-range volume could perform better for moderate distortion levels, indicating a promising direction for our future work.

\subsection{Experiments on KITTI}
On the KITTI test set, comprehensive experiments under asymmetric conditions for numerous methods would exceed the submission limits, potentially violating KITTI's submission constraint~\cite{kitti2012,kitti2015}. Therefore, following~\cite{OpenStereo}, we refer to other methods' leaderboard results under ideal symmetric stereo conditions and only list several reference methods in Table~\ref{tab:kitti}. Despite this handicap, our method achieved comparable or superior performance to many existing algorithms, demonstrating its effectiveness in addressing challenges in asymmetric stereo matching.

\begin{table}[t]
  \centering
  \caption{Ablation study of input processing ($\uparrow$/$\downarrow$: up-/down-sampling). Our proposed method (in bold) performs the best.}\label{tab:input}
\begin{adjustbox}{width=\linewidth}
\begin{tabular}{cccc|c}
\toprule
$I_L$ for $\mathbf{f}^{cor}$ & $I_R$ for $\mathbf{f}^{cor}$ & $I_L$ for $\mathbf{f}^{cat}$ & $I_R$ for $\mathbf{f}^{cat}$ & {EPE} \\ \midrule
-                           & $\uparrow$                  & -                           & $\uparrow$                  & 0.56            \\
$\downarrow$                & -                           & $\downarrow$                & -                           & 1.22            \\
-                           & $\uparrow$                  & $\downarrow$                & -                           &  0.64            \\
$\mathbf{\downarrow}$       & {-}                  & {-}                  & $\mathbf{\uparrow}$         & $\mathbf{0.53}$   \\ \bottomrule
\end{tabular}
\end{adjustbox}
\end{table}

\subsection{Zero-shot Generalization on Middlebury}
The model's generalization ability and robustness in transitioning from simulation to real data are critical for practical applications. Hence, we trained IVF-AStereo and other methods on Scene Flow, and subsequently tested them on Middlebury~2014~\cite{middlebury} in a zero-shot setting. We also assessed their performances under varying degrees of resolution degradation. As shown in Table~\ref{tab:middlebury}, our method demonstrated superior performance in terms of zero-shot sim-to-real generalization and robustness under varying degrees of degradation. Additionally, our mitigation (+A) effectively improved MC-Stereo, which further validates the proposed approach in a zero-shot setting. Figure~\ref{fig:example} visually compares several methods, highlighting the accuracy of our model in overall depth estimation and its superior handling of fine structures that are vulnerable to degradation.

\subsection{Ablation Study}

\subsubsection{Resolution Alignment for Cost Volume}
To align the resolution between $I_L$ and $I_R$ in asymmetric stereo for constructing cost volumes, one can either upsample $I_R$ or downsample $I_L$. Table \ref{tab:input} explores the combinations of resolution alignments. Upsampling $I_R$ for both volumes is straightforward but does not yield optimal performance. Conversely, downsampling $I_L$ for both volumes apparently leads to information loss and suboptimal outcomes. Overall, as explained in the section \ref{sec:volume}, downsampling $I_L$ for $\mathbf{f}^{cor}$ and upsampling $I_R$ for $\mathbf{f}^{cat}$ can achieve the lowest EPE.

\begin{table}[t]
  \centering
    \caption{Ablation study of Multi-peak lookup in the local correlation volume refinement}\label{tab:multipeak}
    \begin{tabular}{c cc}
      \toprule
      Multi-peak lookup & EPE & $>3$px(\%) \\
      \midrule
      $K=1$ (single peak)  & 0.54    & 2.65   \\
      $K=2$   & \underline{0.54}    & \underline{2.65}   \\
      ${K=3}$   & \(\mathbf{0.53} \) & \(\mathbf{2.62} \)  \\
      $K=4$   & 0.54    & 2.70  \\
      
      \bottomrule
    \end{tabular}
\end{table}

\begin{table}[t]
  \centering
    \caption{Ablation study of two-phase fusion scheme in the iterative disparity optimization.}\label{tab:gru-fusion}
  \begin{adjustbox}{width=\linewidth} 
    \begin{tabular}{c c c cc }
      \toprule
      \multirow{3}{*}{Scheme} & \multirow{3}{*}{\( \text{GRU}_1 \rightarrow \text{GRU}_2\)} & \multirow{3}{*}{\( \text{GRU}_2 \rightarrow \text{GRU}_1\)} & \multicolumn{2}{c}{SceneFlow} \\
      \cmidrule(lr){4-5}
      & & & EPE & $>3$px(\%) \\
      \midrule
      1&× & × & 0.57    & 2.83   \\
      2& All iterations & × & 0.54    & 2.71   \\
      3&× & All iterations & 0.54    & 2.62   \\
      4& All iterations & All iterations &  \(\mathbf{0.53} \)    &  \underline{2.62} \\
      \textbf{5}& Phase 2 & All iterations &  \(\mathbf{0.53} \)    &  \(\mathbf{2.59} \)   \\
      \bottomrule
    \end{tabular}
  \end{adjustbox}
\end{table}

\subsubsection{Multi-peak Lookup}
Table \ref{tab:multipeak} examines the effects of varying numbers of groups \( K \) in multi-peak lookup, where the top-$K$ correlations and disparities are selected for updating $\text{GRU}_1$~\cite{mc}. Contrary to MC-Stereo's conclusion, due to information degradation in asymmetric stereo, increasing \( K \) may introduce noise peaks and does not necessarily lead to higher accuracy. Instead, we observe a trend where accuracy initially increases and subsequently decreases, with the best results at \( K = 3 \), which is the configuration we adopt.

\subsubsection{Two-phase Fusion Scheme}
Table \ref{tab:multipeak} evaluates five fusion schemes for iterative disparity optimization with different information flow: 1. no fusion, 2. and 3. one-sided flow from $\text{GRU}_1$ to $\text{GRU}_2$ ($\text{GRU}_1 \rightarrow \text{GRU}_2$) and vise versa ($\text{GRU}_2 \rightarrow \text{GRU}_1$), 4. fusion from the first iteration, and 5. fusion with the proposed two-phase scheme. In the absence of information flow, a zero tensor is used. First, any information flow can improve performance over no fusion. Besides, as $\text{GRU}_2$ exhibits less distortion, $\text{GRU}_2 \rightarrow \text{GRU}_1$ outperforms $\text{GRU}_1 \rightarrow \text{GRU}_2$. Moreover, while fusing two $\text{GRU}$s can increase accuracy, the proposed scheme further enhances the $>3$px by refining initially poor correlation volume (Figure~\ref{fig2}) then fusing appropriately. 

\section{Conclusion and Future Work}
We study the stereo matching with asymmetric visual properties. Based on the observation of cost volume distributions, we first introduce the mitigation for asymmetric stereo that can be compatible with existing methods. Furthermore, we propose an end-to-end learning-based Iterative Volume Fusion network for Asymmetric Stereo matching (IVF-AStereo) to address the challenges in asymmetric stereo. IVF-AStereo achieves superior performance on the Scene Flow dataset and demonstrates state-of-the-art generalization ability on the Middlebury dataset. It is also competitive with existing methods on the KITTI dataset under disadvantaged conditions (ours asymmetric versus others symmetric). 

Though this work primarily focuses on stereo matching, asymmetric stereo can also be promising for image restoration due to the complementary information captured by different views. Additionally, while we mainly tackles the resolution and color asymmetry, we believe the analysis on matching volume distortion can provide insight to more real-world distortions and will be explored in the future. 

\printbibliography

\end{document}